\documentclass{IOS-Book-Article}
\usepackage{amssymb}
\usepackage{amsmath}
\usepackage{amsthm}
\usepackage{graphicx}
\usepackage{tikz}
\usepackage{enumerate}
\usepackage[inline]{enumitem}
\usepackage{multirow}
\usepackage{subcaption}
\usepackage{makecell}
\newtheorem{definition}{Definition}
\newtheorem{example}{Example}
\newtheorem{theorem}{Theorem}

\def\hb{\hbox to 10.7 cm{}}
\newcommand{\key}[1]{\textbf{#1}}
\newcommand{\ikey}[1]{\emph{#1}}
\newcommand{\infer}{\noexpand{| \hspace{-0.5em} \sim}~}
\newcommand{\ninfer}{\noexpand{| \hspace{-0.5em} \not\sim}~}

\newcommand{\short}{\texttt{short}^\pi}
\newcommand{\house}{\texttt{house}^\pi}
\newcommand{\job}{\texttt{job}^\delta}
\newcommand{\bank}{\texttt{bank}^\delta}

\begin{document}

\pagestyle{headings}
\def\thepage{}

\begin{frontmatter}              % The preamble begins here.

%\pretitle{Pretitle}
\title{An Argumentative Explanation Framework for Generalized Reason Model with Inconsistent Precedents\\ (Extended Version)}
%\title{An Argumentative Explanation for Reasoning with Inconsistent Precedents}

%\markboth{}{September 2016\hb}
%\subtitle{Subtitle}

\author[A]{\fnms{Wachara} \snm{Fungwacharakorn}}
\author[B]{\fnms{Gauvain} \snm{Bourgne}}
\author[A]{\fnms{Ken} \snm{Satoh}}

%\runningauthor{B.P. Manager et al.}
\address[A]{Center for Juris-Informatics, ROIS-DS, Tokyo, Japan}
\address[B]{LIP6, Sorbonne University, CNRS, Paris, France}

%, by permitting a court to decide a new case for a side as long as it does not introduce any new inconsistencies

\begin{abstract}
Precedential constraint is one foundation of case-based reasoning in AI and Law. It generally assumes that the underlying set of precedents must be consistent. To relax this assumption, a generalized notion of the reason model has been introduced. While several argumentative explanation approaches exist for reasoning with precedents based on the traditional consistent reason model, there has been no corresponding argumentative explanation method developed for this generalized reasoning framework accommodating inconsistent precedents. To address this question, this paper examines an extension of the derivation state argumentation framework (DSA-framework) to explain the reasoning according to the generalized notion of the reason model.
\end{abstract}

\end{frontmatter}

\section{Background: The Generalized Reason Model}
\label{sec:precedential}

This section provides a background on the reason model of precedential constraint \cite{horty2012factor} and its generalization for reasoning with inconsistent precedents \cite{canavotto2022precedential,canavotto2025reasoning}. The reason model assumes that a dispute has two \ikey{sides}: a \ikey{plaintiff}, denoted as $\pi$, and a \ikey{defendant}, denoted as $\delta$. It also assumes a factor domain $\mathcal{F}$, whose each element is called a \ikey{factor}. Intuitively, a factor represents a relevant fact pattern in a legal decision and is assumed to favor either one side or the other in a dispute. Hence, a fact domain $\mathcal{F}$ can be considered as a partition $\mathcal{F}^\pi = \{f^\pi_1,\ldots,f^\pi_n\}$ (pro-plaintiff factors) and $\mathcal{F}^\delta= \{f^\delta_1,\ldots,f^\delta_m\}$ (pro-defendant factors). A \ikey{reason set} for the side $s \in \{\pi,\delta\}$ is defined as any finite subset of $\mathcal{F}^s$. Meanwhile, a \ikey{fact situation} is defined as any finite subset of $\mathcal{F}$. For any side $s \in \{\pi,\delta\}$ and any fact situation $X$, the opposite side of $s$ is denoted as $\bar{s}$ (i.e., $\pi = \bar{\delta}$ and $\delta = \bar{\pi}$) and the set of all pro-$s$ factors in $X$ is denoted as $X^s$ (i.e., $X^s = X \cap \mathcal{F}^{s}$).

In the reason model, a \ikey{case} is represented as a tuple $\langle X,r,s \rangle$ where $X$ is a fact situation;  $s \in \{\pi,\delta\}$ is the side that the case is decided in favor of (the \ikey{winning} side); and $r$ is a \ikey{rule} of the form $U \rightarrow s$ where $U \subseteq X ^s$ is the reason set to support why this case is decided in favor of side $s$. For any rule $r = U \rightarrow s$, it is defined that $premise(r) = U$ and $conclusion(r)=s$. For any case $c = \langle X,r,s \rangle$, it is defined that $\mathit{facts}(c) = X$, $rule(c) = r$, and $outcome(c) = s$. 

% The reason model considers that a case and a set of cases (called a \ikey{case base}) induce \ikey{a fortiori priorities} over the reason sets. That is, a case $c = \langle X,r,s \rangle$ induces a priority $<_c$ over the reason sets $U \subseteq \mathcal{F}^{\bar{s}}$ and $V \subseteq \mathcal{F}^s$ such that $U <_c V$ if and only if $U \subseteq X$ and $premise(r) \subseteq V$. A case base $\Gamma$ induces a priority order $<_\Gamma$ over the reason sets $U$ and $V$ such that $U <_{\Gamma} V$, if and only if there is a case $c$ in $\Gamma$ such that $U <_c V$.

The reason model considers that a case induces \ikey{a fortiori} priority over reason sets. Specifically, a case $c = \langle X,r,s \rangle$ induces a priority $<_c$ over reason sets for different sides $U \subseteq \mathcal{F}^{\bar{s}}$ and $V \subseteq \mathcal{F}^s$ such that $U <_c V$ if and only if $U \subseteq X^{\bar{s}}$ and $premise(r) \subseteq V$. For the same side, the reason sets with more factors are stronger than those with fewer ones (i.e., for any $s \in \{\pi,\delta\}$, any case = $\langle X,r,s' \rangle$ and any reason sets $U,V \subseteq \mathcal{F}^s, U <_c V$ if and only if $U \subsetneq V$). The reason model also considers that a set of cases, termed as a \ikey{case base}, induces a priority also. A case base $\Gamma$ induces a priority $<_\Gamma$ over reason sets $U$ and $V$ such that $U <_{\Gamma} V$ if and only if there is a case $c$ in $\Gamma$ such that $U <_c V$. Since a case base can contain multiple cases, it can induce a priority such that $U <_{\Gamma} V$ and $V <_{\Gamma} U$ for some reason set $U$ and $V$. Such a pair of $U$ and $V$ is termed an \ikey{inconsistency} in $\Gamma$, denoted as $U \bot_{\Gamma} V$. The set of all inconsistencies in $\Gamma$ is then denoted as $inc(\Gamma)$. If $inc(\Gamma) = \emptyset$, then $\Gamma$ is said to be \ikey{consistent}; otherwise, \ikey{inconsistent}.

The original reason model \cite{horty2012factor} presumes that a case base must be \ikey{consistent} and suggests that the court should maintain the consistency of the case base. That is, given a case base $\Gamma$ representing precedents, the court is allowed to decide the fact situation $X$ in favor of the
side $s$ on the basis of the rule $r$ if and only if $\Gamma \cup \{\langle X, r, s\rangle\}$ is still consistent. Since this principle does not work if the case base is inconsistent in the first place, Canavotto \cite{canavotto2022precedential,canavotto2025reasoning} has generalized the principle so that the court must avoid creating new inconsistencies, rather than maintaining existing consistency. That is, the court is allowed to decide the fact situation $X$ in favor of side $s$ on the basis of the rule $r$ if and only if $inc (\Gamma \cup \{\langle X, r, s\rangle\}) \subseteq inc(\Gamma)$.

For any fact situation $X$, case base $\Gamma$, and side $s \in \{\pi,\delta\}$, the generalized reason model \cite{canavotto2022precedential,canavotto2025reasoning} defines that:
\begin{enumerate}
    \item $\Gamma \infer P_{X}(s)$ holds if and only if there exists a rule $r$ (called a \ikey{permitted} rule) such that $inc (\Gamma \cup \{\langle X, r, s\rangle\}) \subseteq inc(\Gamma)$.
    \item $\Gamma \infer O_{X}(s)$ holds if and only if all permitted rule $r$ has a conclusion $s$; that is , $\Gamma \infer P_{X}(s)$ and $\Gamma \ninfer P_{X}(\bar{s})$.
\end{enumerate}

This is connected to deontic logic. Sometimes, the court is \ikey{permitted} (denoted as $P_X$) to decide a fact situation in favor of any side as it does not create any new inconsistencies, but sometimes the court is \ikey{obligated} (denoted as $O_x$) to decide in favor of one side as it cannot decide in favor of the other side without creating new inconsistencies. Formally, it has been observed that, for any fact situation $X$ and case base $\Gamma$, exactly one of the following holds (adapted from Observation 4 in \cite{canavotto2022precedential}):
    \begin{itemize}
        \item $\Gamma \infer P_{X}(\pi)$ and $\Gamma \infer P_{X}(\delta)$
        \item $\Gamma \infer O_{X}(\pi)$ -- that is, $\Gamma \infer P_{X}(\pi)$ and $\Gamma \ninfer P_{X}(\delta)$
        \item $\Gamma \infer O_{X}(\delta)$ -- that is, $\Gamma \infer P_{X}(\delta)$ and $\Gamma \ninfer P_{X}(\pi)$
    \end{itemize}

This follows from the observation that the court can always decide any fact situation in favor of one side at least. That is, it is never the case that $\Gamma \ninfer P_{X}(\pi)$ and $\Gamma \ninfer P_{X}(\delta)$ (adapted from Observation 3 in \cite{canavotto2022precedential}). Example \ref{ex:tax} is adapted from Example 1 in \cite{canavotto2022precedential}, used to illustrate the generalized reason model.

\begin{example}[fiscal domicile]
\label{ex:tax}
This example considers disputes about whether a person can change their income tax address (fiscal domicile) after working aboard for a while. We have two sides, the side of the home country of the person as the plaintiff ($\pi$), which is arguing against change of fiscal domicile; and the side of the person as the defendant ($\delta$), which is arguing for change of fiscal domicile. We consider the following factors (with the favoring side in the superscript):

\begin{enumerate}
     \item $\short$: the defendant spent only one month in the foreign country
    \item $\house$:  the defendant still owned a house in the home country
    \item $\job$: 
    the defendant had a permanent job in the foreign country
    \item $\bank$: 
    the defendant opened a bank account in the foreign country
\end{enumerate}
In this example, we assume a case base $\Gamma_1 = \{c_1,c_2\}$ with two precedent cases:
\begin{enumerate}
    \item $c_1 = \langle \{\short,\job\}, \{\short\} \rightarrow \pi, \pi\rangle$: the defendant spent only one month, but already had a permanent job in the foreign country. The case has been decided in favor of the plaintiff ($\pi$) for the reason that the defendant had spent only one month in the foreign country.
    \item $c_2 = \langle \{\short,\job,\bank\}, \{\job\} \rightarrow \delta, \delta\rangle$: the defendant spent only one month, but already had a permanent job and a bank account in the foreign country. The case has been decided in favor of the defendant ($\delta$) for the reason that the defendant had a permanent job.
\end{enumerate}

Suppose that we have a new fact situation $X_1 = \{\short,\house,\job\}$: the defendant spent only one month and still owed a house in the home country, but already had a permanent job in the foreign country. The question is which side the court should decide in favor of.
\end{example}

\begin{figure}[h]
    \centering
    \begin{subfigure}{0.48\textwidth}
        \begin{tikzpicture}[>=latex,line join=bevel]
              \node(ns) at (0bp,0bp) { $\{\short\}$};
              \node(nj) at (100bp,0bp) {$\{\job\}$};
              \node(nsh) at (0bp,60bp) {$\{\short,\house\}$};
                     
              \draw[->] (nj)--(ns) node[midway,above] {$c_1$};
              \draw[->] ([yshift=-5]ns.east)--([yshift=-5]nj.west) node[midway,below] {$c_2$};
              \draw[->] (ns)--(nsh);
              \draw[blue,->] (nj)--(nsh) node[midway,right] {$\langle X_1, \_, \pi \rangle$};
        \end{tikzpicture}
        \caption{with $\langle X_1, \_, \pi \rangle$}
        \label{fig:diagram-1}
    \end{subfigure}
    \hfill
    \begin{subfigure}{0.48\textwidth}
        \begin{tikzpicture}[>=latex,line join=bevel]
              \node(ns) at (0bp,0bp) { $\{\short\}$};
              \node(nj) at (100bp,0bp) {$\{\job\}$};
              \node(nsh) at (0bp,60bp) {$\{\short,\house\}$};
                     
              \draw[->] (nj)--(ns) node[midway,above] {$c_1$};
              \draw[->] ([yshift=-5]ns.east)--([yshift=-5]nj.west) node[midway,below] {$c_2$};
              \draw[->] (ns)--(nsh);
              \draw[red,dashed,->] (nsh)--(nj) node[midway,right] {$\langle X_1, \_, \delta \rangle$};
        \end{tikzpicture}
        \caption{with $\langle X_1, \_, \delta \rangle$}
        \label{fig:diagram-2}
    \end{subfigure}
    \caption{Diagrams depicting priorities induced by $\Gamma_1$ with auxiliary cases from $X_1$}
    \label{fig:diagram}
\end{figure}
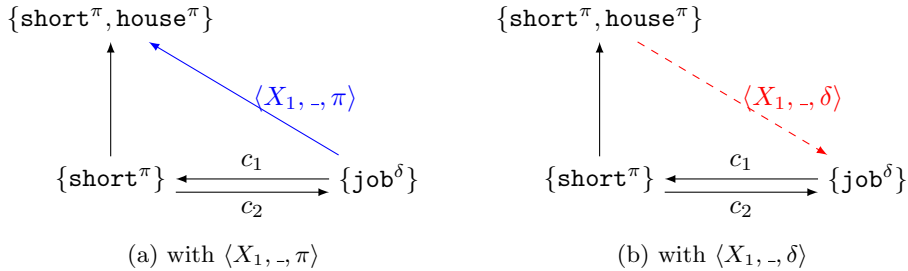

Figure \ref{fig:diagram} shows the diagrams that depict the priorities induced by $\Gamma_1 = \{c_1, c_2\}$ in Example \ref{ex:tax}. Meanwhile, the diagram also reflects the priorities over reason sets for different sides, which are induced by the cases $c_1$ and $c_2$. Furthermore, we can see an inconsistency as a cycle in the diagram. 
For instance, $\Gamma_1$ is inconsistent since $\{\job\} <_{c_1} \{\short\}$ and $\{\short\} <_{c_2} \{\job\}$; hence, $\{\short\} \bot_{\Gamma_1} \{\job\}$. Now, considering the fact situation $X_1 = \{\short,\house,\job\}$ in Example \ref{ex:tax}. If the court decides $X_1$ in favor of $\pi$, then a new case $\langle X_1,\_,\pi \rangle$ induces a new priority $\{\job\} < \{\short,\house\}$ (we note the rule as $\_$ because this is true regardless of the rule in the new case). The new priority does not create additional inconsistencies, as no new cycles are created in Figure \ref{fig:diagram-1}. However, if the court decides $X_1$ in favor of $\delta$, then a new case $\langle X_1,\_,\delta \rangle$ induces a new priority $\{\short,\house\} < \{\job\}$, which creates a new inconsistency $\{\short, \house\} \bot_{\Gamma_1} \{\job\}$. We can also see this inconsistency derived from a new cycle $\{\short\} - \{\short, \house\} - \{\job\}$ in Figure \ref{fig:diagram-2}. Therefore, the court is obligated to decide $X_1$ in favor of $\pi$. That is, $\Gamma_1 \infer O_{X_1}(\pi)$. 

\section{Background: Abstract Argumentation Framework and Dispute Tree}
\label{sec:aa-framework}

Since DSA-framework is extended from abstract argumentation framework \cite{dung1995acceptability} and the framework provides explanations using dispute trees \cite{dung2007computing}, this section recaps some concepts discussed in those works as follows.
\ikey{Abstract argumentation framework} (AA-framework) is a pair $(\mathcal{A}, \mathcal{R})$. Each element of $\mathcal{A}$ represents an argument and $\mathcal{R} \subseteq \mathcal{A} \times \mathcal{A}$. For $x,y \in \mathcal{A}$, we say $x$ attacks $y$ if and only if $x \mathcal{R} y$. For any AA-framework $(\mathcal{A}, \mathcal{R})$, $E \subseteq \mathcal{A}$, and $x,y \in \mathcal{A}$, it is said that: 
\begin{enumerate}
    \item $E$ \emph{attacks} $x$ if some argument $z \in E$ attacks $x$;
    \item $E$ \emph{defends} $y$ if, for every $x \in \mathcal{A}$ that attack $y$, $E$ attacks $x$.
    \item $E$ is \emph{conflict-free} if no $x,y \in E$ such that $x$ attacks $y$;
    \item $E$ is \emph{admissible} if $E$ is conflict-free and $E$ defends every $z \in E$;
    \item $E$ is the \emph{grounded extension} of the AA-framework if it can be constructed inductively as $E = \bigcup_{i \geq 0} E_i$, where $E_0$ is the set of unattacked arguments, and $\forall i \geq 0$, $E_{i+1}$ is the set of arguments that $E_{i}$ defends;
    \item $E$ is a \emph{stable extension} of the AA-framework if it is a conflict-free set that attacks every argument that does not belong in $E$;
    \item $E$ is a \emph{preferred extension} of the AA-framework if it is a maximal admissible set with respect to the set inclusion;
    \item $E$ is a \emph{complete extension} of the AA-framework if it is an admissible set and every argument that $E$ defends, belongs to $E$.
\end{enumerate}

Meanwhile, a dispute tree \cite{dung2007computing} is considered as a tree which is a subgraph of AA-framework. The definition has been extended in several papers, but in this paper, we extend the definition from abstract argumentation for case-based reasoning \cite{cyras2016abstract}. For any AA-framework,  a \emph{dispute tree} for an argument $x_0 \in \mathcal{A}$, is a (possibly infinite) tree $\mathcal{T}$ such that:
\begin{enumerate}
    \item every node of $\mathcal{T}$ is of the form $[L:x]$, with $L \in \{P, O\}$ and $x \in \mathcal{A}$ where $L$ indicates the status of proponent ($P$) or opponent ($O$);
    
    \item the root of $\mathcal{T}$ is $[P:x_0]$;
    
    \item for every proponent node $[P:y]$ in $\mathcal{T}$ and for every $x \in \mathcal{A}$ such that $x$ attacks $y$, there exists $[O:x]$ as a child of $[P:y]$;
    
    \item for every opponent node $[O:y]$ in $\mathcal{T}$, there exists at most one child of $[P:x]$ such that $x$ attacks $y$;
    
    \item there are no other nodes in $\mathcal{T}$ except those given by 1-4.
    \end{enumerate}

A dispute tree $\mathcal{T}$ is admissible if and only if
\begin{enumerate*}[label=(\alph*)]
    \item every opponent node $[O:x]$ in $\mathcal{T}$ has a child, and
    \item no $[P:x]$ and $[O:y]$ in $\mathcal{T}$ such that $x=y$.
\end{enumerate*} 

\section{DSA-Framework for Generalized Reason Model}
\label{sec:dsa-framework}

This section presents an extension of the derivation state argumentation framework (DSA-framework) for explaining the generalized reason model. DSA-framework was originally defined to explain preemption in hierarchical constraint represented norms \cite{fungwacharakorn_vecomp_2024}. The framework extends abstract argumentation framework by structuring arguments called \key{derivation state arguments} (DS-arguments). Basically, DS-arguments consider maximal structures and derivation states with respect to each subset of fact situation, called a \key{situational knowledge}. In the setting of precedential constraint, we consider \ikey{conclusiveness} and a \key{derivation state} as follows.
%  (if the subset is a proper subset, it is called a \ikey{partial} situational knowledge; otherwise, a \ikey{complete} situational knowledge)

\begin{definition}[conclusiveness and derivation state]
\label{def:maximal-consistent}
Let $X$ be a fact situation and $\Gamma$ be a case base.  $\Gamma$ is said to be \ikey{conclusive} with respect to $X$ if and only if there exists $s \in \{\pi,\delta\}$ such that $X^{\bar{s}} <_{\Gamma} X^s$ and $X^s \not<_{\Gamma} X^{\bar{s}}$; then, $s$ is said to be a \ikey{derivation state} for $X$ with respect to $\Gamma$.
\end{definition}

Intuitively, a derivation state indicates for which side the court is obligated to decide a fact situation in accordance to the original reason model. This makes an inconsistent case base $\Gamma$ with $X^s \bot_{\Gamma} X^{\bar{s}}$, for any fact situation $X$, not conclusive with respect to $X$. In that case, we consider a maximal subset of case base that is conclusive, called a \key{maximal conclusive sub-base}, defined as follows.

\begin{definition}[maximal conclusive sub-base]
A \ikey{maximal conclusive sub-base} of $\Gamma$ with respect to $X$ is a $\supseteq$-maximal subset of $\Gamma$ that is conclusive with respect to $X$. That is, $\gamma$ is a subset of $\Gamma$, conclusive with respect to $X$, and no $\gamma' \subseteq \Gamma$ that is conclusive with respect to $X$ such that $\gamma \subsetneq \gamma'$. The set of all maximal conclusive sub-bases of $\Gamma$ with respect to $X$ is denoted by $max(\Gamma,X)$.
\end{definition}

For any fact situation $X$ and case base $\Gamma$, there are at most two maximal conclusive sub-bases of $\Gamma$ with respect to $X$ (i.e., $||max(\Gamma,X)|| \leq 2$). The setting in which two maximal conclusive sub-bases occur is when $X^\pi \bot_\Gamma X^\delta$. In that setting, one maximal conclusive sub-base has a derivation state $\pi$ and another has a derivation state $\delta$. Now, we define DS-arguments to be used in DSA-framework as follows.

\begin{definition}[DS-argument]
\label{def:ds-argument}
Let $X$ be a fact situation and $\Gamma$ be a case base. A \ikey{derivation state argument} (DS-argument) with respect to $X$ and $\Gamma$ is any tuple $(\chi,\gamma,s) \in 2^X \times 2^\Gamma  \times \{\pi,\delta\}$ such that 
\begin{enumerate*}[label=(\alph*)]
    \item $\chi \subseteq X$;
    \item  $\gamma \in max(\Gamma,\chi)$; and
    \item $s$ is a derivation state for $\chi$ with respect to $\gamma$.
\end{enumerate*}
The set of all DS-arguments with respect to $X$ and $\Gamma$ is denoted as $\mathcal{D}_{X,\Gamma}$.
\end{definition}

The following theorem links the property of DS-arguments with obligations described in the generalized reason model.

\begin{theorem}
\label{th:complete-situation}
  For any fact situation $X$, case base $\Gamma$, and $s \in \{\pi,\delta\}$, $\Gamma \infer O_{X}(s)$ if and only if 
  \begin{itemize}
      \item there does not exist $\gamma \subseteq \Gamma$ such that $(X,\gamma,\bar{s}) \in \mathcal{D}_{X,\Gamma}$; and
      \item there exists $\gamma' \subseteq \Gamma$ such that $(X,\gamma',s) \in \mathcal{D}_{X,\Gamma}$.
  \end{itemize}
\end{theorem}

\begin{proof}
    (From left to right)  $\Gamma \infer O_{X}(s)$ implies that $\Gamma \ninfer P_{X}(\bar{s})$. Then, there is no case $\langle X, r, \bar{s}\rangle$ that can be added to $\Gamma$ without creating new inconsistencies. Thus, there is no maximal conclusive sub-base $\gamma \subseteq \Gamma$ with respect to $X$ for the derivation state $\bar{s}$; otherwise, there exists $c \in \gamma$ such that $X^s <_c X^{\bar{s}}$ and $\langle X, rule(c), \bar{s}\rangle$ can be added to $\Gamma$ without creating new inconsistencies. Meanwhile, we must have $X^{\bar{s}} <_{\Gamma} X^s$ and hence there must exist $\gamma' \subseteq \Gamma$ such that $(X,\gamma',s) \in \mathcal{D}_{X,\Gamma}$; otherwise, we can add $\langle X, X^{\bar{s}} \rightarrow\bar{s},\bar{s}\rangle$ to $\Gamma$ without creating new consistencies, contradicting $\Gamma \ninfer P_{X}(\bar{s})$.
    
    (From right to left) If there does not exist $\gamma \subseteq \Gamma$ such that $(X,\gamma,\bar{s}) \in \mathcal{D}_{X,\Gamma}$, then there is no conclusive subset of $\Gamma$ with respect to $X$ for the derivation state $\bar{s}$ and hence $X^s \not<_{\Gamma} X^{\bar{s}}$. Thus, we can add a case $\langle X, X^s \rightarrow s,s\rangle$, without creating new consistencies so 
    $\Gamma \infer P_{X}(s)$. Meanwhile, 
    there exists $\gamma' \subseteq \Gamma$ such that $(X,\gamma',s) \in \mathcal{D}_{X,\Gamma}$. It implies that $X^{\bar{s}} <_{\gamma'} X^s$ and hence $X^{\bar{s}} <_{\Gamma} X^s$, meaning that no case $c = \langle X, r, \bar{s}\rangle$ that can be added to $\Gamma$ without creating $X^s <_c X^{\bar{s}}$, which leads to new consistencies so $\Gamma \ninfer P_{X}(\bar{s})$. Therefore, $\Gamma \infer O_{X}(s)$.
\end{proof}

Next, we extend DSA-framework for explaining the generalized reason model. Intuitively, DSA-framework introduces attacks between DS-arguments based on the change of derivation state due to the expansion of situational knowledge. The extended framework is formally defined as follows.

\begin{definition}[DSA-framework for explaining generalized reason model]
  Let $X$ be a fact situation and $\Gamma$ be a case base. DSA-framework with respect to $X$ and $\Gamma$ is $(\mathcal{D}_{X,\Gamma},\mathcal{R})$ satisfying the following conditions.
  \begin{enumerate}
      \item $\mathcal{D}_{X,\Gamma}$ is the set of all DS-arguments with respect to $X$ and $\Gamma$.
      \item For $( \chi,\gamma,s ), ( \chi',\gamma',s' ) \in \mathcal{D}_{X,\Gamma}$,  $( \chi,\gamma,s )$ attacks $( \chi',\gamma',s' )$ if and only if
      \begin{itemize}
          \item (change derivation state) $s \neq s'$, and 
          \item (gain more knowledge) $\chi' \subsetneq \chi$, and 
          \item (concise attack)  no $(\chi'',\gamma'',s)$ in $\mathcal{D}_{X,\Gamma}$ with $\chi' \subsetneq \chi'' \subsetneq \chi$. 
      \end{itemize}
  \end{enumerate}
\end{definition}

Although the definition of attacks does not derive from the maximal conclusive sub-bases used in the arguments, there are some properties related to those sub-bases. For example, attacking arguments cannot have larger maximal conclusive sub-bases (i.e., if $(\chi,\gamma,s)$ attacks $(\chi',\gamma',s')$ then $\gamma \not\supset \gamma'$). This is because a maximal conclusive sub-base with respect to $\chi$ is also conclusive with respect to $\chi'$ situations, so if $\gamma \supsetneq \gamma'$ then $\gamma'$ becomes not maximal.

Since attacks in DSA-framework are derived from subset relations, DSA-framework is well-founded, meaning that the framework is free from attack cycles. Following the property of abstract argumentation \cite{dung1995acceptability}, the extension of well-founded framework is unique. That is, there is only one extension that is grounded, stable, preferred, and complete. 

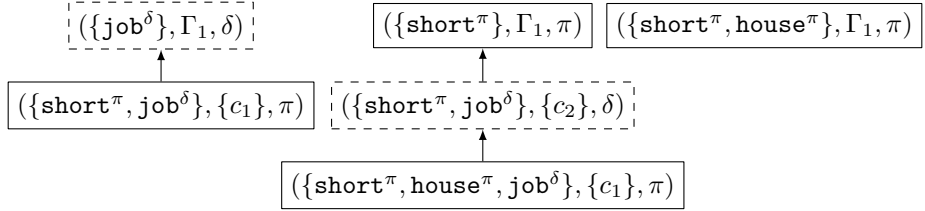
\begin{figure}[h]
    \begin{tikzpicture}[>=latex,line join=bevel]
      \node (n11) at (-25bp,80bp) [draw,dashed] {$(\{\job\}, \Gamma_1, \delta)$};
      \node (n12) at (-25bp,50bp) [draw] {$(\{\short,\job\}, \{c_1\}, \pi)$};
      
      \node (n21) at (95bp,80bp) [draw] {$(\{\short\}, \Gamma_1, \pi)$};
      \node (n22) at (95bp,50bp) [draw,dashed] {$(\{\short,\job\}, \{c_2\},\delta)$};
      \node (n23) at (95bp,20bp) [draw] {$(\{\short,\house,\job\}, \{c_1\},\pi)$};

      \node (n31) at (200bp,80bp) [draw] {$(\{\short,\house\}, \Gamma_1,\pi)$};
    
      \draw [->] (n12)--(n11);
      \draw [->] (n22)--(n21);
      \draw [->] (n23)--(n22);
    \end{tikzpicture}
    \caption{DSA-framework from Example \ref{ex:tax} (Arguments with solid frames are inside the extension while arguments with dashed frames are outside the extension)}
    \label{fig:dsa-framework}
\end{figure}

Figure \ref{fig:dsa-framework} shows DSA-framework with respect to the situation $X_1$ and the case base $\Gamma_1$ in Example \ref{ex:tax}. In the framework, there are two DS-arguments with situational knowledge $\{\short,\job\}$ because there are two maximal conclusive sub-bases $\{c_1\}$ and $\{c_2\}$ with respect to the situational knowledge, following the inconsistency $\{\short\} \bot_{\Gamma_1} \{\job\}$. Furthermore, we can see that all and only arguments with derivation state $\pi$ are included in the extension. This is a consequence from that court is obligated to decide $X_1$ in favor of $\pi$. The link between extension and obligation can be demonstrated in the following theorem.

\begin{theorem}
\label{th:extension}
  For any fact situation $X$, case base $\Gamma$, and $s \in \{\pi,\delta\}$, $\Gamma \infer O_{X}(s)$ if and only if the extension of DSA-framework $(\mathcal{D}_{X,\Gamma},\mathcal{R})$ is not empty and includes all and only DS-arguments with derivation state $s$. 
\end{theorem}

\begin{proof}
    Let $E$ be the extension of $(\mathcal{D}_{X,\Gamma},\mathcal{R})$, which is grounded, stable, preferred, and complete.  
    
    (From left to right) It is obvious that if $\Gamma \infer O_{X}(s)$, then $E$ is not empty. Thus, we prove by contradiction that if non-empty $E$ does not include all and only DS-arguments with derivation state $s$, there must be some $(\chi_j, \gamma_j, \bar{s}) \in \mathcal{D}_{X,\Gamma}$ that is unattacked.
    \begin{itemize}
        \item Suppose, toward contradiction, that $(\chi_1,\gamma_1,\bar{s}) \in E$, there must be some $(\chi_2,\gamma_2,s)$ that attacks $(\chi_1,\gamma_1,\bar{s})$ and there must be $(\chi_3,\gamma_3,\bar{s}) \in E$ that attacks $(\chi_2,\gamma_2,s)$ because $E$ is admissible. Inductively, we have that there must be some $(\chi_j,\gamma_j,\bar{s}) \in E$ that is unattacked. 
        \item Suppose, toward contradiction, that $(\chi_4,\gamma_4,s) \in \mathcal{D}_{X,\Gamma} \setminus E$, there must be some $(\chi_5,\gamma_5,\bar{s}) \in E$ that attacks $(\chi_4,\gamma_4,s)$ because $E$ is stable. There are two possible cases \begin{enumerate*}[label=(\alph*)]
            \item $(\chi_5,\gamma_5,\bar{s})$ is unattacked; or 
            \item $(\chi_5,\gamma_5,\bar{s})$ is attacked by $(\chi_6,\gamma_6,s) \in \mathcal{D}_{X,\Gamma}$ which is not in $E$ because $E$ is conflict-free; inductively, there must be some $(\chi_j, \gamma_j, \bar{s}) \in E$ that is unattacked.
        \end{enumerate*}
    \end{itemize}
    The existence of unattacked $(\chi_j, \gamma_j, \bar{s}) \in E$ contradicts Theorem \ref{th:complete-situation} that there does not exist $\gamma \subseteq \Gamma$ such that $(X,\gamma,\bar{s}) \in \mathcal{D}_{X,\Gamma}$ and $(X,\gamma',s) \in \mathcal{D}_{X,\Gamma}$ because such $(X,\gamma',s)$ would attack $(\chi_j,\gamma_j,\bar{s})$ if it existed. 
    
    (From right to left) If $E$ is not empty and includes all and only DS-arguments with derivation state $s$, then there exists  $(\chi_i,\gamma',s) \in \mathcal{D}_{X,\Gamma}$ which is unattacked because $(\mathcal{D}_{X,\Gamma},\mathcal{R})$ is well-founded and $E$ is the grounded extension. It also implies that $(X,\gamma',s) \in \mathcal{D}_{X,\Gamma}$ and 
    there does not exist $\gamma \subseteq \Gamma$ such that $(X,\gamma,\bar{s}) \in \mathcal{D}_{X,\Gamma}$. Therefore, $\Gamma \infer O_X(s)$ according to Theorem \ref{th:complete-situation}.
\end{proof}

Also, in Figure \ref{fig:dsa-framework}, we can see that the framework can be decomposed into three disconnected trees. These trees correspond to dispute trees described in Section \ref{sec:aa-framework}, which we introduce as explanations for why the court is obligated to decide a fact situation in favor of one side. The explanations are formally defined as follows. 

%Next, we introduce dispute trees built from DSA-framework as explanations. In Figure \ref{fig:dsa-framework}, the framework can be decomposed into three disconnected trees, each of which corresponds to a dispute tree described in Section \ref{sec:aa-framework}. Following this, we define \key{explanations} for why the court is obligated to decide a fact situation in favor of one side as follows. 

\begin{definition}[explanation]
    Let $X$ be a fact situation, $\Gamma$ be a case base, and $s \in \{\pi,\delta\}$. An \ikey{explanation} for why the court is obligated to decide $X$ in favor of side $s$ by following $\Gamma$ is any admissible dispute tree for any non-attacking DS-arguments (i.e., DS-arguments that do not attack any other DS-arguments in DSA-framework $(\mathcal{D}_{X,\Gamma},\mathcal{R})$) with derivation state $s$, and we denote the set of all of such explanations as $\mathcal{E}_{\Gamma,X}(s)$.
\end{definition}

The following theorem proves that
if the court is obligated to decide a fact situation in favor of one side, there is always an explanation for why it is and no explanation for why it is not.

\begin{theorem}
\label{th:explanation}
    For any fact situation $X$, case base $\Gamma$, and $s \in \{\pi,\delta\}$, $\Gamma \infer O_{X}(s)$ if and only if $\mathcal{E}_{\Gamma,X}(s) \neq \emptyset$ and $\mathcal{E}_{\Gamma,X}(\bar{s}) = \emptyset$. 
\end{theorem}

\begin{proof}
 This follows from the proof that there is an admissible dispute tree for every argument inside the extension (see \cite{dung2007computing,cyras2016explanation}). (From left to right) According to Theorem \ref{th:extension}, there always exist DS-arguments with derivation state $s$ inside the extension.  Therefore, $\mathcal{E}_{\Gamma,X}(s) \neq \emptyset$. Meanwhile, there does not exist an admissible dispute tree for every argument outside the extension. Therefore, $\mathcal{E}_{\Gamma,X}(\bar{s}) = \emptyset$. (From right to left) If $\mathcal{E}_{\Gamma,X}(s) \neq \emptyset$ and $\mathcal{E}_{\Gamma,X}(\bar{s}) = \emptyset$, all and only DS-arguments with derivation state $s$ are included in the extension. Therefore, $\Gamma \infer O_X(s)$ according to Theorem \ref{th:extension}. 
\end{proof}

From Figure \ref{fig:dsa-framework},  three decomposed trees can be viewed as three different trees. For the tree on the left, the root proponent node proposes that the court is obligated to decide $\{\job\}$ in favor of $\delta$. Then, the opponent node challenges that the court is obligated to decide the situation with incremental knowledge $\{\short, \job\}$ in favor of $\pi$. Because the proponent cannot defend with more incremental situational knowledge with respect to $X_1 = \{\short,\house,\job\}$, this tree is not admissible and hence not an explanation. For the tree in the middle, the root proponent node proposes that the court is obligated to decide $\{\short\}$ in favor of $\pi$. Then, the opponent node challenges that the court is obligated to decide the fact situation with incremental knowledge $\{\short, \job\}$ in favor of $\delta$, which can be defended by the proponent that the court is obligated to decide the fact situation with more incremental knowledge $\{\short, \house, \job\}$ in favor of $\pi$. Hence, the tree in the middle is admissible and hence an explanation. For the tree on the right, the root proponent node proposes that the court is obligated to decide $\{\short,\house\}$ in favor of $\pi$. This node cannot be challenged, so the tree is admissible and hence an explanation. The explanations guide that adding the factor $\house$ is the main reason why the court is obligated to decide in favor of $\pi$. That is because adding $\house$ reverses the favor side from $\delta$ to $\pi$ (explained by the tree in the middle). In other words, we can explain that the court is obligated to decide the situational knowledge $\{\short,\house\}$ in favor of $\pi$ and the obligation cannot be challenged by other situational knowledge of $X_1$ (explained by the tree on the right).

\section{Concluding Remark}
\label{sec:concluding}

The paper has presented an argumentative explanation approach for explaining the generalized reason model of precedential constraint for inconsistent precedents. The approach is based on the extension of the derivation state argumentation framework (DSA-framework), in which each argument considers a maximal conclusive subset of case-base corresponding to each subset of fact situation, and attacks are derived from the change of derivation state due to the expansion of situational knowledge. This makes DSA-framework one of argumentative approaches to explain behaviors of normative systems by considering such explanations. Aligning with other similar argumentative approaches (e.g., \cite{wyner2024satisfaction,rotolo2025rule}), the framework has a similar property, such that attacking arguments cannot have larger maximal conclusive sub-bases. It implies that, in the setting of inconsistent case-bases,  we sometimes trade off cases in the conclusive sub-base in order to move the argument to larger fact situations; and, to achieve that, hierarchies between cases are implicitly introduced. For example, Figure \ref{fig:dsa-framework} demonstrates that $\{c_2\}$ prevails over $\{c_1\}$ as we need to worsen $\{c_2\}$ into $\{c_1\}$ to expand the situational knowledge from $\{\short,\job\}$ to $\{\short,\house,\job\}$ but we never worsen $\{c_1\}$ into $\{c_2\}$ in the same fact situation. Future research can explore such dynamic hierarchies, which may be relevant to static hierarchical factors introduced in hierarchical precedential constraint \cite{van2023hierarchical,canavotto2023hierarchies}.

Meanwhile, since this paper focuses on the setting of precedential constraint, the proposed framework inherits some properties that are different from those explanation frameworks for other normative systems. For example, because the reason model of precedential constraint introduces a side as a conclusion to a rule, it does not lead to an ambiguity as in normative constraint hierarchies, in which a rule can be specified with several conclusions (see \cite{fungwacharakorn2025normative}). The proposed framework also contains only two derivation states, making it easier to understand the framework compared to three or more of them. Several challenges mentioned in \cite{canavotto2025reasoning} remain unanswered here but we can foresee some potentials. For example, this paper has not investigated an explanation for two -side permissions (i.e., $\Gamma \infer P_x(s)$ and $\Gamma \infer P_x(\bar{s})$) but it may be possible by introducing parallel explanations (i.e., $\mathcal{E}_{\Gamma,X}(s) \neq \emptyset$ and $\mathcal{E}_{\Gamma,X}(\bar{s}) \neq \emptyset$) or rebutting attacks (two-way attacks) between arguments with the same situational knowledge but different derivation states, e.g., ($\{\short,\job\},\{c_1\},\pi)$ and $\{\short,\job\},\{c_2\},\delta)$ in Figure \ref{fig:dsa-framework}. Rebuttal attacks clearly reflect inconsistencies, but they allow a rebutting argument to defend itself, making infinite and possibly inconsistent dispute trees. Therefore, further refinements of the framework can be investigated to address this.

\section*{Acknowledgements}
This work was supported by the “R\&D Hub Aimed at Ensuring Transparency
and Reliability of Generative AI Models” project of the Ministry of
Education, Culture, Sports, Science and Technology and JSPS KAKENHI
Grant Numbers, JP22H00543, JP25H00522,  JP25H01112 and JP25H01152.

%\bibliographystyle{ios1}
%\bibliography{mybib}

\end{document}